\documentclass[11pt, a4paper]{article}
\usepackage{latexsym}
\usepackage[fleqn]{amsmath}
\usepackage{graphicx}
\usepackage[fleqn]{amsmath}
\usepackage[varg]{txfonts}
\usepackage[english]{babel}
\usepackage[hyperindex=false,
    pageanchor=false,
    pdfmenubar=false,
    pdfpagelabels=false, breaklinks=true]{hyperref}
\usepackage{natbib}
\bibpunct{(}{)}{,}{a}{,}{,}

\usepackage{setspace}
\usepackage{epstopdf}
\usepackage{setspace}
\usepackage{graphicx}
\usepackage{threeparttable}
\newcommand{\argmin}{\operatornamewithlimits{argmin}}
\newcommand{\mb}{\mathbb} % Must use capital letter \mb{A} not \mb{a}
\newcommand{\vc}{\mathbf} % use smale letter\mbf{a}

\newcommand{\mc}{\mathcal}
\newcommand{\mr}{\mathrm}

\begin{document}
%\title{ Influence function of  Mean Element, Covariance operator and  Kernel Canonical Correlation Analysis}         
\title{ Gene-Gene association for Imaging Genetics Data using Robust Kernel Canonical Correlation Analysis} %and its Influence Function}% for Imaging Genetics Analysis}   
\author{\textbf{  Md. Ashad Alam$^{1,2}$ and Osamu Komori$^{3}$ and Yu-Ping Wang$^1$}  \\ $^{1}$Department of Biomedical Engineering, Tulane University\\
 New Orleans, LA 70118, USA\\
$^2$Department of Statistics, Hajee Mohammad Danesh Science and Technology\\ University, Dinajpur 5200, Bangladesh\\       %Enter your name between curly 
$^3$Department of Electrical and Electronics Engineering, University of Fukui,Fukui 910-8507, Japan}

\date{}
\maketitle
\begin{abstract}
In genome-wide interaction studies, to detect gene-gene interactions, most methods are divided into two folds: single nucleotide polymorphisms (SNP) based and gene-based methods. Basically, the methods based on  the gene are more effective than the methods based on a single SNP. Recent years,  while the  kernel canonical correlation analysis (Classical kernel CCA) based U statistic (KCCU) has proposed to detect the  nonlinear relationship between genes. To estimate the variance in KCCU, they  have used resampling based methods which are highly computationally intensive. In addition, classical kernel CCA is not robust  to  contaminated data. We, therefore,  first discuss robust kernel mean element, the robust kernel covariance, and cross-covariance operators. Second, we propose a method based on influence function  to  estimate the variance of the KCCU.  Third, we propose a nonparametric robust KCCU  method based on robust kernel CCA, which is designed for contaminated data and  less sensitive to noise than classical kernel CCA. Finally, we investigate the proposed methods to synthesized  data and imaging genetic data set.  Based on gene ontology and pathway analysis, the synthesized and  genetics analysis demonstrate that the proposed  robust method shows the superior performance of the  state-of-the-art methods. 
\end{abstract}  
%\keywords{Robustness, Kernel CCA, Robust kernel CCA,  Gene-gene interaction, Imaging genetic data}

\section{Introduction}
\label{sec:Intro}
Due to a large number of human single nucleotide polymorphisms (SNPs), kernel methods, methods using the positive definite kernel, have become a popular and effective tool for conducting  genome-wide association studies (GWAS), especially for identifying disease associated genes.  They offer real-world and principled algorithms to learn how a large number of genetic variants are associated with complex phenotypes, to help expose the complexity in the relationship between the genetic markers and the outcome of interest.  Human complex diseases are usually caused by the combined effect of multiple genes without any standard patterns of inheritance.  Indeed, to gain a better understanding of the genetic mechanisms and explain the pathogenesis of human complex diseases, the detection of interactions between genes (GGIs) is important instead of SNP-SNP interactions (SSIs). SSI methods, which examine one SNP in a gene, cannot completely interpret the GGIs. Conversely, GGI methods, which consider genes that have many SNPs,  not only take into account the interactions between SNPs but also the interactions between genes  \citep{Wang-14,Li-15}.

In the last decade, a number of statistical methods have been used to detect  GGIs. Logistic regression, multifactor dimensionality reduction, linkages disequilibrium and entropy based statistics are examples of such methods. Among others, whole genome association analysis toolsets such as  \citep[minPtest]{Hieke-14},\citep[BOOST]{Wan-10}, \citep[BEAM]{Zhan-07},\citep[Random Jungle]{Schwarz-10},\citep[Tuning ReliefF]{Moore-07}, and \citep[PLINK]{Purcell-07}, have also been developed by the  genomics, bioinformatics and biomedical communities.  While most of these methods are based on the unit association of SNP, testing the associations between the phenotype and SNPs has limitations and is not sufficient for interpretation of GGIs \citep{Zhongshang-12}. A powerful tool for multivariate gene-based genome-wide associations is proposed \citep[MGAS]{Sluis-15}. In the case-control study, linear canonical correlation based U statistic (CCU) is  utilized  to identify  gene-gene interaction \citep{Peng-10}. In recent years, this method was extended to nonlinear  statistics using  kernel canonical correlation analysis (Classical kernel CCA), which is proposed in \citep{Akaho}. Extending linear CCU to the reproducing kernel Hilbert space (RKHS) is known as kernel CCU (or KCCU)\citep{Larson-2014,Li-12}. To estimate the variance in KCCU, researchers have used resampling-based methods, despite their high computational burden.

Bootstrapping, a resampling method that takes a sample and resamples several times, can be prohibitively expensive for large data or a complicated estimator. It may also have poor finite sample performance. Fortunately, the influence function (IF), the effect of change in a single observation of an estimator, directly relates to the asymptotic distribution of the estimator. As such, IF is a convenient way to find the variances and covariances of a variety of estimators\citep{Huber-09,Hampel-11}.

Classical kernel CCA, weighted multiple kernel CCA and other kernel methods have been extensively studied in unsupervised kernel fusion for decades \citep{Yu-11, Ge-15}. But these methods  are not robust; they are sensitive to contaminated data \citep{Ashad-08, Ashad-10}. Including classical kernel CCA, the unsupervised methods explicitly or implicitly depend on kernel mean elements (kernel ME), the kernel covariance operator (kernel CO) and/or kernel cross-covariance operator (kernel CCO). They can be formulated as an empirical optimization problem to achieve robustness by combining empirical optimization problem with ides of Huber or Hamples M-estimation model. The robust kernel CO and CCO can be computed efficiently via kernelized iteratively re-weighted least square (KIRWLS) problem \citep{Ashad-16}.  Even though a number  of researchers have investigated the robustness issue for supervised learning, especially the support vector  machine for classification and regression \citep{Christmann-04,Christmann-07,Debruyne-08}, there are no general well-founded robust methods for unsupervised learning.

Robustness is an essential and challenging  issue in statistical machine learning for  multiple source data analysis because  outliers, data  that cause surprise in relation to the majority of the data,  often occur in the real data. Outliers may be correct, but we need to examine them for transcription errors. They can cause havoc with classical statistical or statistical machine learning methods. To overcome this problem, since 1960 many robust methods have been developed, which are  less sensitive to outliers.  The goals of robust statistics are to use statistical methods on the whole dataset and to identify points deviating from the original pattern for further investigation \citep{Huber-09,Hampel-11}. In recent years, a robust kernel density estimation (robust kernel DE) has been proposed \citep{Kim-12}, which is less sensitive than the kernel density estimation. Through an empirical comparison and   sensitivity analysis, it has been shown that classical kernel CCA is as sensitive to outliers as kernel PCA \citep{Ashad-10,Ashad-14T}.

The contribution of this paper is fourfold. First, we address the robust kernel ME,  the robust kernel CO, and  CCO. Second, we propose a method based on IFs  to  estimate the variance of the CCU. Third, we propose a nonparametric robust CCU  method based on robust kernel CCA, which is designed for contaminated data and  less sensitive to noise than classical kernel canonical correlation analysis. Finally we apply the proposed methods to synthesized  data and imaging genetic data sets. Experiments on synthesized (both ideal data (ID) and contaminated data (CD))  and  genetics analysis demonstrate that the proposed  robust method performs markedly better than the state-of-the-art methods.

This paper is organized as follows. In the next section, we provide a brief review of the robust kernel ME, the robust kernel CO, the robust kernel CCO and the robust Gram matrices.  In Section $3$,  we discuss in  brief   the classical  kernel CCA, robust kernel CCA and KCCU. After a brief review of classical  kernel CCA in  Section \ref{sec:CKCCA}, we propose the  robust kernel CCA in   Section \ref{sec:RKCCA} and  the IF based test statistic to estimate the variance of the CCU is proposed in \ref{sec:IFT}. In Section \ref{sec:exp}, we describe  experiments conducted on both synthesized data and  the imaging genetics data sets. We conclude with a summary of findings and mention areas of future research in Section \ref{sec:con}.  

\section{Robust kernel (cross-) covariance operator}
As shown in \citep{Kim-12,Ashad-16,Ashad-16a}    the   kernel ME is the   solution to the empirical risk optimization problem, which is the least square  type  estimator.  
\begin{eqnarray}
\label{EROP1}
\argmin_{f\in \mc{H}_X} \frac{1}{n}\sum_{i=1}^n\| \vc{\Phi}(X_i)- f\|^2_{\mc{H}_X}.
\end{eqnarray} 
 As  shown in   Eq. (\ref{EROP1}), we can define kernel CCO  as  an empirical risk optimization problem. Given the  pair of  independent and identically distributed sample, $(X_i, Y_i)_{i=1}^n$,  the kernel CCO is an operator of the RKHS, $\mc{H}_X\otimes\mc{H}_Y$,
\begin{eqnarray}
\label{EROP2}
\argmin_{\Sigma_{XY}\in \mc{H}_X\otimes \mc{H}_Y}\frac{1}{n} \sum_{i=1}^n \| \vc{\Phi}_c(X_i) \otimes \vc{\Phi}_c(Y_i) - \Sigma_{XY}\|^2_{\mc{H}_X\otimes\mc{H}_Y},
\end{eqnarray} 
where $\vc{\Phi}_c(X_i)= \vc{\Phi}(X_i)-\frac{1}{n}\sum_{b=1}^n \vc{\Phi}(X_b)$. In the special case that Y is equal to X, we get the kernel CO.

This type of   estimator is sensitive  to the presence of  outliers in the features. In recent years, the robust kernel ME has been proposed for density estimation  \citep{Kim-12}. Our goal is to extend this notion to kernel CO and kernel CCO. To do so, we estimate kernel CO and kernel CCO   based on the robust loss functions, M-estimator. The estimated kernel CO and kernel CCO are called robust kernel CO and robust kernel CCO, respectively. The most common example of robust loss functions, $\zeta(t)$ on $t \geq 0$, are  Huber's and Hampel's loss functions.  Unlike the quadratic loss function, the derivatives of these loss functions are bounded \citep{Huber-09, Hampel-86}. Huber's  loss function  is  defined as
\begin{eqnarray}
\zeta(t)=
\begin{cases}
	 t^2/2\qquad  \qquad ,0\leq t\leq c  
\\
 ct-c^2/2\qquad  ,c\leq t \nonumber 
\end{cases}
\end{eqnarray}
and Hampel's loss function is defined as
\begin{eqnarray}
\zeta(t)=
\begin{cases}
	 t^2/2\qquad  \qquad\qquad\qquad ,0\leq t\le c_1  
\\
 c_1t-c_1^2/2\qquad\qquad  ,c_1\leq t < c_2
\\
c_1(t-c_3)^2/2 (c_2-c_3)+ c_1(c_2+c_3-c_1)/2  ,c_2\leq t < c_3\\
 c_1(c_2+c_3-c_1)/2 \qquad \qquad ,c_3\leq t. \nonumber
\end{cases}
\end{eqnarray}

Given the weights $\vc{w}=[ w_1, w_2, \cdots, w_n]^T$ of the robust kernel ME of a set of observations,   the points $\vc{\Phi}_c(X_i):= \vc{\Phi}(X_i) - \sum_{a=1}^nw_a \vc{\Phi}(X_a)$  are centered and the centered Gram matrix is $\tilde{K}_{ij}=(\vc{H}\vc{K}\vc{H}^T)_{ij}$,  
where $\vc{1}_n=[1, 1, \cdots, 1]^T$ and $\vc{H}=\vc{I}- \vc{1}_n\vc{w}^T$.
 Eq. (\ref{EROP2}) can be written as    
\begin{eqnarray}
\label{REROP1}
\argmin_{\hat{\Sigma}_{XY}\in {\mc{H}_X\otimes\mc{H}_Y}} \frac{1}{n}\sum_{i=1}^n \zeta(\| \vc{\Phi}_c(X_i) \otimes \vc{\Phi}_c(Y_i) - \Sigma_{XY}\|_ {\mc{H}_X\otimes\mc{H}_Y}).
\end{eqnarray}
As shown in \citep{Kim-12},  Eq. (\ref{REROP1})  does not have a closed form solution, but   using the kernel trick the classical re-weighted least squares (IRWLS) can be extended to a RKHS. The solution is then,
\[\widehat{\Sigma}_{XY}^{(h)}= \sum_{i=1}^n w_i^{(h-1)}\tilde{k}(X, X_i)\tilde{k}(Y, Y_i),\]
where $w_i^{(h)}=\frac{\varphi(\|\vc{\Phi}_c(X_i)\otimes \vc{\Phi}_c(Y_i) - \Sigma_{XY}\|_{\mc{H}_X\otimes\mc{H}_Y})}{\sum_{b=1}^n\varphi(\| \vc{\Phi}_c(X_b)\otimes\vc{\Phi}_c(Y_b)- \Sigma_{XY}\|_{\mc{H}_X\otimes\mc{H}_Y})}\,, \rm{and} \, \varphi(x)=\frac{\zeta^\prime(x)}{x}.$

Given weight of robust kernel mean element \\$\vc{w}=[w_1, w_2, \cdots, w_n]^T$ of a set of observations $X_1, \cdots, X_n$, the points 
\begin{eqnarray}
\vc{\Phi}_c(X_i):= \vc{\Phi}(X_i)- \sum_{b=1}^nw_b \vc{\Phi}(X_b) \nonumber
\end{eqnarray} 
 are centered. Thus
\begin{eqnarray}
\tilde{K}_{ij}=\langle \vc{\Phi}_c(X_i), \vc{\Phi}_c(X_j)\rangle &=& ( (\vc{I}- \vc{1}_n\vc{w}^T) \vc{K} (\vc{I}- \vc{1}_n \vc{w}^T)^T)_{ij} \nonumber\\
&=&(\vc{H}\vc{K}\vc{H}^T)_{ij},
\label{CM1}
\end{eqnarray}
where $\vc{1}_n=[1, 1, \cdots, 1]^T$ and $\vc{H}=\vc{I}- \vc{1}_n\vc{w}^T$. For a set of test points $ X^t_1, X_2^t, \cdots, X_T^t$,  we define two matrices of order $T\times n$ as  $K_{ij}^{test}= \langle \Phi(X^t_i), \Phi(X_j) \rangle$ and  $\tilde{K}_{ij}^{test}= \langle \Phi(X^t_i)- \sum_{b=1}^nw_b \Phi(X_b),  \Phi(X_j)- \sum_{d=1}^nw_d  \Phi(\vc{X}_b)\rangle$
As in Eq. (\ref{CM1}), the robust centered Gram matrix of test points,  $K_{ij}^{test}$,  in terms of the robust Gram matrix is defined as,
\begin{eqnarray}
\label{CM2}
\tilde{K}_{ij}^{test}=  K_{ij}^{test}- \vc{1}_T \vc{w}^T \vc{K}- \vc{K}^{test} \vc{w}\vc{1}_n^T+ \vc{1}_T \vc{w}^T \vc{K} \vc{w}\vc{1}^T_n \nonumber
\end{eqnarray}
The algorithms of estimating  robust kernel CC and CCO  are discussed in \citep{Ashad-16}.

\section{Classical kernel CCA and Robust kernel CCA}
Classical kernel CCA has been proposed as a nonlinear extension of linear CCA \citep{Akaho,Lai-00}. This method along with its variant have  been applied for various purposes including  genomics, computer graphics and computer-aided drug discovery and  computational biology \citep{Alzate2008,Ashad-14T,Ashad-13}.  Theoretical results on the convergence of kernel CCA have also been  obtained  \citep{Fukumizu-SCKCCA,Hardoon2009}.
\subsection{Classical kernel CCA}
\label{sec:CKCCA}
The aim of classical kernel CCA  is to seek  the sets of  functions in the RKHS for which the correlation (Corr) of  random variables  is maximized. The simplest case, given  two sets of random variables $X$  and $Y$ with  two   functions in the RKHS, $f_{X}(\cdot)\in \mc{H}_X$  and  $f_{Y}(\cdot)\in \mc{H}_Y$, the optimization problem of  the random variables $f_X(X)$ and $f_Y(Y)$ is
\begin{eqnarray}
\label{ckcca1}
\max_{\substack{f_{X}\in \mc{H}_X,f_{Y}\in \mc{H}_Y \\ f_{X}\ne 0,\,f_{Y}\ne 0}}\mr{Corr}(f_X(X),f_Y(Y)),
\end{eqnarray}
where the functions $f_{X}(\cdot)$ and $f_{Y}(\cdotp)$ are obtained up to scale.

We can extract  the desired functions with a  finite sample. Given an i.i.d sample, $(X_i,Y_i)_{i=1}^n$ from a  joint distribution $F_{XY}$, by taking the inner products with elements or ``parameters" in the RKHS, we have  features
$f_X(\cdot)=\langle f_X, \Phi_X(X)\rangle_{\mc{H}_X}= \sum_{i=1}^na_X^ik_X(\cdot,X_i) $ and
 $f_Y(\cdot)=\sum_{i=1}^na_Y^ik_Y(\cdot,Y_i)$, where $k_X(\cdot, X)$ and $k_Y(\cdot, Y)$ are the associated kernel functions for $\mc{H}_X$ and $\mc{H}_Y$, respectively. The kernel Gram matrices are defined as   $\vc{K}_X:=(k_X(X_i,X_j))_{i,j=1}^n $ and $\vc{K}_Y:=(k_Y(Y_i,Y_j))_{i,j=1}^n $.  We need the centered kernel Gram matrices $\vc{M}_X=\vc{C}\vc{K}_X\vc{C}$ and $\vc{M}_Y=\vc{C}\vc{K}_Y\vc{C}$, where $ \vc{C} = \vc{I}_n -\frac{1}{n}\vc{B}_n$ with  $\vc{B}_n = \vc{1}_n\vc{1}^T_n$ and $\vc{1}_n$ is the vector with $n$ ones.
The empirical estimate of Eq. (\ref{ckcca1}) is based on

\begin{align*}
 %\label{ckcca6}
& \widehat{\rm{Cov}}(f_X(X),f_Y(Y))
= \frac{1}{n} \vc{a}_X^T\vc{M}_X\vc{M}_Y \vc{a}_Y= \vc{a}_X^T\vc{M}_X\vc{W}\vc{M}_Y \vc{a}_Y , \\
& \widehat{\rm{Var}}( f_X(X))
=\frac{1}{n} \vc{a}_X^T\vc{M}_X^2 \vc{a}_X= \vc{a}_X^T\vc{M}_X \vc{W} \vc{M}_X \vc{a}_X, \,  \\ &\widehat{\rm{Var}}( f_Y(Y))=\frac{1}{n} \vc{a}_Y^T\vc{M}_Y^2 \vc{a}_Y= \vc{a}_Y^T\vc{M}_Y \vc{W} \vc{M}_Y\vc{a}_Y, 
\end{align*}
where  $\vc{W}$ is a diagonal matrix with elements  $\frac{1}{n}$, and  $\vc{a}_{X}$ and $\vc{a}_{Y}$ are the  directions of   $X$ and $Y$, respectively. The regularized coefficient $\kappa > 0$.

\subsection{Robust kernel CCA}
\label{sec:RKCCA}
In this section, we propose  a  robust kernel CCA method based on the robust kernel CO and the robust kernel CCO. While many robust linear CCA  methods have been proposed that fit the bulk of the data well and indicate the points deviating from the original pattern  for further investment \citep{Adrover-15, Ashad-10}, there are no general well-founded robust methods of  kernel CCA. The  classical kernel CCA  considers the same weights for each data point, $\frac{1}{n}$,  to estimate kernel CO and kernel CCO, which is the solution of an empirical risk optimization problem using the quadratic loss function. It is known that the least square loss function is not a robust loss function. Instead, we can solve  an empirical risk optimization problem using the robust least square loss function and  the weights are  determined based on data via KIRWLS. The robust kernel CO and kernel CCO are used in classical kernel CCA, which we call a robust kernel CCA method. Figure \ref{RobustKCCA} presents  a detailed algorithm  of the proposed method (except for the first two steps, all steps are similar to classical kernel CCA). This method  is designed for  contaminated data as well, and  the principles we describe apply also to the  kernel methods, which must deal with the issue of kernel CO and kernel CCO.

\begin{figure}
\hrule
%\vspace*{3mm}
Input: $D=\{(X_1,Y_1), (X_2,Y_2), \ldots, (X_n,Y_n) \}$ in $\mb{R}^{m_1\times m_2}$. 
\begin{enumerate}
\item Calculate the robust cross-covariance operator, $\hat{\Sigma}_{YX}$ using algorithm as in \citep{Ashad-16}.
\item Calculate the robust covariance operator $\hat{\Sigma}_{XX}$ and $\hat{\Sigma}_{YY}$ using the same weight of the cross-covariance operator (for simplicity).
\item Find $ \mb{B}_{YX} = (\hat{\Sigma}_{YY} + \kappa \vc{I})^{- \frac{1}{2}} \hat{\Sigma}_{YX} (\hat{\Sigma}_{XX} + \kappa\vc{I})^{- \frac{1}{2}}$
\item For $\kappa > 0$, we have $\rho^2_j$,  the largest eigenvalue of $\mb{B}_{YX}$ for $j=1, 2,\cdots, n$.
\item  The unit  eigenfunctions of   $\mb{B}_{YX}$  corresponding to the  $j$th eigenvalues  are $\hat{\xi}_{jX}\in \mc{H}_X$ and $\hat{\xi}{j_Y}\in \mc{H}_Y$ 
\item The \it {j}th ($j= 1, 2, \cdots, n$)  kernel canonical variates are given by
\[ \hat{f}_{jX}(X) = \langle \hat{f}_{jX}, \tilde{k}_X(\cdot, X)\rangle \,\rm{and}\,\hat{f}_{jY}(X) = \langle \hat{f}_{jY}, \tilde{k}_Y(\cdot, Y) \rangle\]
 \rm{where}   $\hat{f}_{jX} =  (\hat{\Sigma}_{XX} + \kappa\vc{I})^{-\frac{1}{2}}\hat{\xi}_{jX}$ \rm{and}   $f_{jY} =  (\hat{\Sigma}_{YY} + \kappa\vc{I})^{-\frac{1}{2}}\hat{\xi}_{jY}$
\end{enumerate}
Output:  the robust kernel CCA
%\vspace*{3mm}
\hrule
\caption{The algorithm of estimating robust  kernel CCA}
\label{RobustKCCA}
\end{figure}

\subsection{Test Statistic}
\label{sec:IFT}
Given the data matrix  $(X_{ij}^{g}, X_{ij^\prime}^{g^\prime})_{i=1}^{D}$   
 with the gene set $g > g^\prime \in\{1, 2, \cdots, G\}$, the number of SNP in each gene, $j \in \{ 1, 2, \cdots, S_g\}$ $j^\prime \in \{ 1, 2, \cdots, S_{g^\prime}\}$ with $D$ case data and similarly  $(X_{ij}^{g}, X_{ij^\prime}^{g^\prime})_{i=1}^{C}$ is for  $C$ control data. Now we apply kernel CCA on the case data and the control data, and  the first kernel canonical correlation is noted as $r^D_{g,g^\prime}$ and $r^C_{g,g^\prime}$, respectively. We can also use the same procedure for the other data sets, for example DNA methylation  and fMRI.  

For correlation test statistics, we need to use the Fisher variance stabilizing transformation of the kernel CC,  defined  as
\[ z^D_{g,g^\prime}= \frac{1}{2}\left[log(1+r^D_{g,g^\prime}r)-log(1-r^D_{g,g^\prime})\right],\] and 
\[ z^C_{g,g^\prime}= \frac{1}{2}\left[log(1+r^C_{g,g^\prime}r)-log(1-r^C_{g,g^\prime})\right],\]
which are approximately distributed as standard normal. To assess the statistical significance  for each pair of genes $g$ and $g^\prime$, we determine the co-association between case and controls. The nonlinear test statistic is define as
\begin{eqnarray}
\label{KUUC}
T_{g,g^\prime}=\frac{ z^D_{g,g^\prime}- z^C_{g,g^\prime}}{\sqrt{\rm{var}(z^D_{g,g^\prime})+\rm{var}(z^D_{g,g^\prime})}},
\end{eqnarray}
which is asymptotically distributed as a standard normal distribution, $N(0,1)$ under the null hypothesis, $z^D_{g,g^\prime}=z^C_{g,g^\prime}$ with independent case and control.
 
As discussed in Section \ref{sec:Intro}, the Bootstrapping methods can be prohibitively expensive for large data or a complicated estimator and also have poor finite sample performance. Fortunately, IFs are directly related to the asymptotic distribution of the estimator, thus using IFs is a convenient way to find the variances and covariances of a variety of estimators.

In this paper, we apply a method  based on IF of kernel CCA, proposed in \citep{Ashad-16},  to  estimate the  variance of the test statistics in Eq. (\ref{KUUC}). To do this, we need to relate the IF of kernel CC to the IF of  Fisher variance stabilizing transformation. Fortunately, the IF of Fisher's transform of the correlation coefficient, $\rho$, is independent of $\rho$ \citep{Devlin-75} and the IF of  Fisher's transform, $z(\rho)$ has the distribution of a product of two independent standard normal variables.

Given $(X_i, Y_i)_{i=1}^n$ be a sample from the joint distribution $F_{XY}$, the empirical IF of first  kernel canonical correlation (kernel CC) at $(X^\prime, Y^\prime)$ is  defined as 
\begin{multline}
\label{EIFKCCA}
\rm{EIF} (X^\prime, Y^\prime, \hat{\rho}^2)= - \hat{\rho}^2 \bar{f}_{X}^2(X^\prime) + 2 \hat{\rho} \bar{f}_{X}(X^\prime) \bar{f}_{Y}(Y^\prime)  - \hat{\rho}^2 \bar{f}_{Y}^2(Y^\prime).
\end{multline}
Letting $x$ and $y$  the standardized sum of and difference between centered  kernel canonical vectors (kernel CV), $\bar{f}_{X}^2(X^\prime)$ and $\bar{f}_{Y}^2(X^\prime)$, respectively. The  EIF  $z(\rho)$  with $u = \frac{(x+y)}{\sqrt{2}}$ and $v = \frac{(x-y)}{\sqrt{2}}$ is then defined as 
\begin{multline}
\label{EIFFT}
\rm{EIF} (X_i, Y_i, z(\hat{\rho}^2))= u_iv_i, \qquad i=1, 2, \cdots, n.
\end{multline}
According to \citep{Hampel-11,Huber-09,Mark-01}, the variance of Fisher's transform is $\rm{var}z(\hat{\rho})= \frac{1}{n}\rm{var}(\rm{EIF} (X, Y, z(\hat{\rho}^2)))= \frac{1}{n^2} \sum_{i=1}^nu_iv_i$. As shown in Section \ref{sec:exp}, the time complexity of this estimator is  lower than the resampling based estimators for instance bootstrap method. 

We can define a similar test statistic for the robust kernel CCA using the  robust kernel CC and the robust kernel CV.

\section{Experiments}
\label{sec:exp}
 We demonstrate the experiments on synthesized and imaging genetics analysis. For the synthesized experiment, we generate two types of  data original data  and those with $5\%$ of  contamination, which is called ideal data (ID) and  contaminated data (CD), respectively.  In all experiments, for the  bandwidth of Gaussian kernel we use  the median of the  pairwise distance \citep{Gretton-08,Sun-07}. Since the goal is to seek the outliers observation, the regularized parameter of  kernel CCA  is set to   $\kappa = 10^{-5}$. The description of the  synthetic data sets  and  the real data sets  are in Section \ref{Sec:syd}  and  Sections \ref{Sec:RD}, respectively.

\subsection{Synthetic data}
\label{Sec:syd}
We conduct simulation studies to evaluate the performance of the proposed methods with the following synthetic data.
\textbf {Sine and cosine function structural (SCS) data:}
 We used a uniform   marginal distribution, and transformed the data with two periodic $\sin$ and $\cos$ functions to make two sets, $X$ and $Y$, respectively, with additive Gaussian noise:
$Z_i\sim U[-3\pi,3\pi], \,\eta_i\sim N(0,10^{-2}),~\,i=1,2,\ldots, n,
 X_{ij}=\sin(jZ_i)+\eta_i,\,  Y_{ij} = \cos(jZ_i)+\eta_i, j=1,2,\ldots, 100.$
For the CD $\eta_i\sim N(1,10^{-2})$.

\textbf {Multivariate Gaussian structural (MGS) data: }
Given  multivariate normal data, $\vc{Z}_i\in\mb{R}^{12} \sim \vc{N}(\vc{0},\Sigma)$ ($i= 1, 2, \ldots, n$) where  $\Sigma$ is the same as in \citep{Ashad-15}. We   divided $\vc{Z}_i$ into two sets of variables ($\vc{Z}_{i1}$,$\vc{Z}_{i2}$), and used the first six variables of $\vc{Z}_i$ as $X$ and perform $\log$  transformation of the absolute value of the remaining  variables ($\log_e(|\vc{Z}_{i2}|)$) as $Y$. For the CD  $\vc{Z}_i\in\mb{R}^{12} \sim \vc{N}(\vc{1},\Sigma)$ ($i= 1,2,\ldots, n$).

\textbf {SNP and fMRI structural (SMS) data:}
Two data sets of SNP data X with $1000$ SNPs and fMRI data Y with 1000 voxels were simulated. To correlate the SNPs with the voxels, a  latent model is used   as in \citep{Parkhomenko-09}. For contamination,  we consider the signal level, $0.5$  and  noise level, $1$  to $10$ and $20$, respectively.

In the  synthetic experiments, first, we investigate  asymptotic relative  efficiency (ARE) of bootstrap based variance  and influence function (IF) based variance for  linear CCA and classical kernel CCA using SCS data with the different sample sizes $n\{100, 200, 300, 400, 500, 750, 1000\}$.  We repeat the experiment  with 100 samples  for each sample size.  To illustrate the computational cost, we also mention the CPU time (in seconds) of each estimator. The configuration of the computer is  an Intel (R) Core (TM) i7 CPU 920@ 2.67 GHz.,  12.00 GB of memory and a 64-bit operating system. Table \ref{tb1:ARETIME} shows the ARE values and times. This table clearly indicates  that the variance based on the IF is  highly efficient for sample size $n > 100$  of  kernel methods and  for the linear CCA   $n <300$. The bootstrap based variance estimates have very high time complexity.

\begin{table*}
 \begin{center}
%\scalebox{0.55}[0.6]{
\caption {The asymptotic relative efficiency of bootstrap (Boot) and influence function (IF) based variance estimation and time (in second) of linear canonical analysis (LCCA) and kernel canonical analysis (KCCA)}
\label{tb1:ARETIME}
 \begin{tabular}{ccccccc} \hline
&\multicolumn{2}{c}{\rm{ARE}}&\multicolumn{4}{c}{\rm{ Time}}\\ 
n&\rm{LCCA}&\rm{KCCA}&\multicolumn{2}{c}{\rm{LCCA}}&\multicolumn{2}{c}{\rm{KCCA}}\\\cline{4-7}
&&&\rm{Boot}&\rm{IF}&\rm{Boot}&\rm{IF}\tabularnewline  \hline
$100$& $5.7079\pm 3.0479$&$0.6181\pm 0.3893$&$0.03$&$0.01$&$0.45$&$0.17$\tabularnewline  
$200$& $2.314\pm2.1520$&$1.1127\pm 0.6306 $&$0.04$&$0.01$&$2.34$&$1.09$\tabularnewline  
$300$& $0.6725\pm0.3519$&$1.2563\pm 0.8776$&$0.04$&$0.03$&$7.02$&$3.47$\tabularnewline 
$400$& $0.5903\pm0.4695 $&$1.3825\pm0.9811 $&$0.04$&$0.02$&$17.16$&$9.50$\tabularnewline  
$500$& $0.4816\pm0.3166 $&$2.2916\pm 1.9775 $&$0.04$&$0.02$&$27.97$&$13.81$\tabularnewline  
$750$& $0.4181\pm 0.2579$&$7.9847\pm 4.3284$&$0.05$&$0.03$&$119.43$&$66.94$\tabularnewline  
$1000$& $0.3814\pm0.2308$&$10.9049\pm5.0561 $&$0.05$&$0.03$&$280.49$&$151.71$\tabularnewline  \hline
\end{tabular}
%}
\end{center}
\end{table*}
We evaluate the performance of  the proposed methods,  robust kernel CCA, in three different settings. The robust kernel CCA compares  with the classical kernel CCA using Gaussian kernel (same bandwidth and regularization).  We consider the same EIF  as shown in Eq (\ref{EIFKCCA}) for both methods. To measure the influence, we calculate the  ratio between  ID and CD of IF of kernel CC and kernel CV.   Based on this ratio, we define two  measures  on kernel  CC and kernel CV,  
\begin{eqnarray}
\eta_{\rho} &=&\left | 1- \frac{\|EIF(\cdot, \rho^2) ^{ID}\|_F}{\|EIF(\cdot, \rho^2)^{CD}\|_F}\right| \qquad \rm{and} \\ 
\eta_{f} &=& \left| 1- \frac{\|EIF(\cdot, f_X)^{ID}- EIF(\cdot,f_Y)^{ID}\|_F}{\|EIF(\cdot, f_X) ^{CD}-EIF(\cdot, f_Y)^{CD}\|_F}\right|,\nonumber
\end{eqnarray}
respectively. The method, which does not depend on the contaminated data, the above measures, $\eta_{\rho}$  and $\eta_{f}$, should  be approximately zero. In other words, the best methods should give small values. To compare, we consider 3 simulated data sets:  MGSD, SCSD, SMSD with $3$ sample sizes, $n\in \{ 100, 500, 1000\}$. For each sample size, we repeat the experiment for $100$ samples.  Table \ref{tbl:ifnorm} presents the results (mean $\pm$ standard deviation) of classical kernel CCA and robust kernel CCA. From this table, we observe that robust kernel CCA outperforms the classical kernel CCA in all cases.

\begin{figure*}
\begin{center}
\includegraphics[width=18cm, height=12cm]{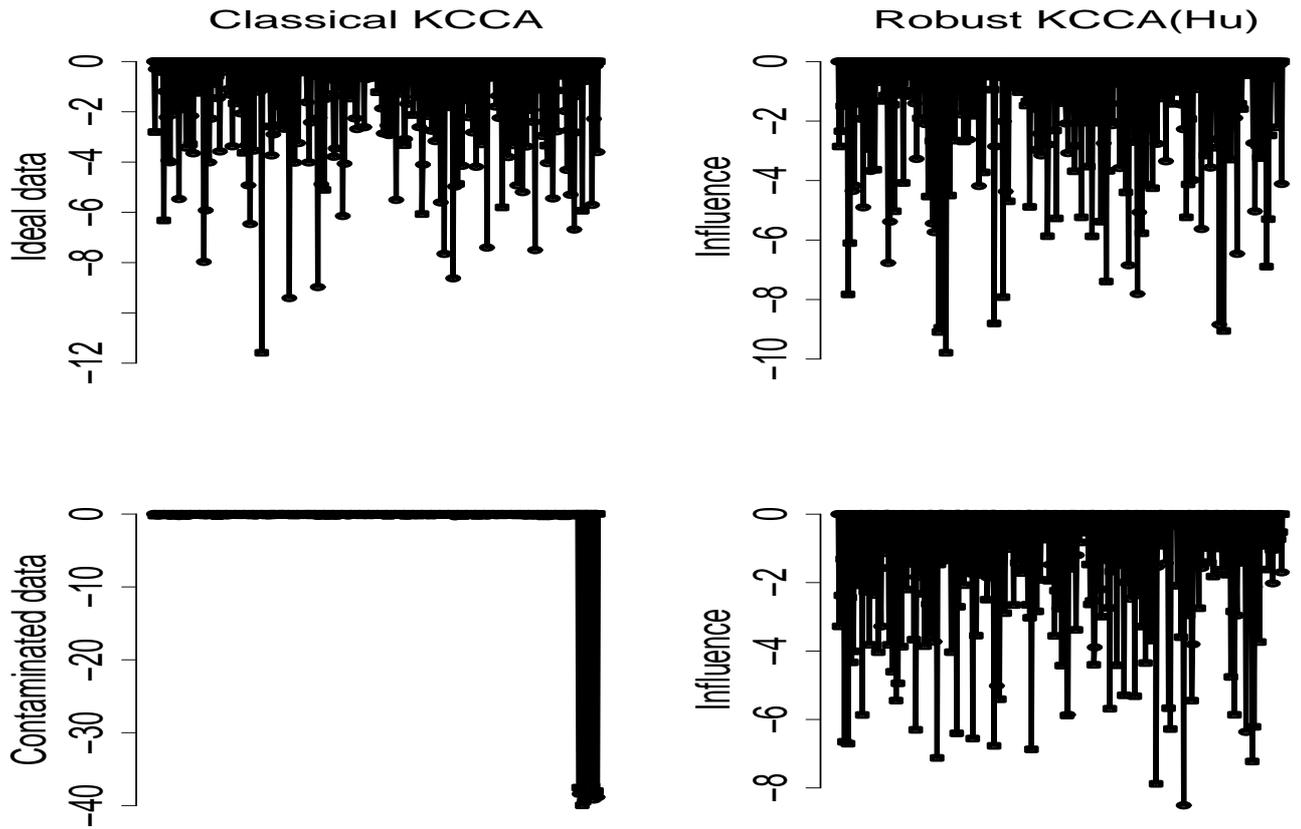}
\caption{The influence  points of SMDS   for ideal  and  contaminated data using classical and robust kernel CCA.}
\label{SMDSIFOB}
\end{center}
\end{figure*}

\begin{table*}
 \begin{center}
\caption {Mean and standard deviation of  measures, $\eta_{\rho}$ and $\eta_{f}$ of classical kernel CCA (Classical) and  robust kernel CCA (Robust).}
\label{tbl:ifnorm}
%\scalebox{0.55}[0.6]{
 \begin{tabular}{llcccccccc} \hline
&\rm{Measure} &\multicolumn{2}{c}{$\eta_{\rho}$}&\multicolumn{2}{c}{$\eta_{f}$}\\ \cline{3-6}
\rm{Data}&$n$&$\rm{Classical}$&$\rm{Robust}$&$\rm{Classical}$&$\rm{Robust}$  \\ \hline
&$100$&$1.9114\pm 3.5945$&$ 1.2445\pm 3.1262$&$ 1.3379\pm 3.5092$&$ 1.3043\pm 2.1842$\tabularnewline
MGSD&$500$&$ 1.1365\pm 1.9545$&$ 1.0864\pm 1.5963$&$ 0.8631 \pm 1.3324 $&$0.7096\pm 0.7463$\tabularnewline
&$1000$&$ 1.1695\pm 1.6264$&$ 1.0831\pm 1.8842$&$ 0.6193 \pm 0.7838$&$ 0.5886\pm 0.6212$\\ \hline
&$100$&$0.4945\pm 0.5750$& $0.3963\pm 0.4642$& $1.6855\pm 2.1862$& $0.9953\pm 1.3497$\tabularnewline
SCSD &$500$&$0.2581\pm 0.2101$& $0.2786\pm 0.4315$& $1.3933\pm 1.9546$&$ 1.1606\pm 1.3400$\tabularnewline
&$1000$&$0.1537\pm 0.1272$&$ 0.1501\pm 0.1252$&$ 1.6822\pm 2.2284$&$1.2715\pm 1.7100$\\ \hline
&$100$&$ 0.6455 \pm 0.0532$& $ 0.1485\pm 0.1020$& $ 0.6507 \pm 0.2589 $& $2.6174\pm 3.3295$ \tabularnewline
SMSD&$500$&$0.6449\pm 0.0223$& $0.0551\pm 0.0463$&$ 3.7345 \pm 2.2394$& $ 1.3733 \pm 1.3765$\tabularnewline
&$1000$&$ 0.6425 \pm 0.0134$& $ 0.0350\pm 0.0312$& $ 7.7497\pm 1.2857$& $ 0.3811 \pm 0.3846$\\ \hline
\end{tabular}
%}
\end{center}
\end{table*}
By the simple graphical display, the index plots (the observations on $x$-axis and the influence on $y$-axis)  assess the related influence data points in data fusion with respect to EIF based on kernel CCA and robust kernel CCA \citep{Ashad-16a}. To do this, we consider SMS Data.  Figure \ref{SMDSIFOB} shows the index plots of  classical kernel CCA and robust kernel CCA. The $1$st and $2$nd rows, and columns of this figure are  for ID and CD, and classical kernel CCA (Classical KCCA) and robust kernel CCA (Robust KCCA), respectively.  These plots show that both methods have almost similar results to  the ID. However, it is clear that the classical kernel CCA is affected by the CD significantly. We can easily identify influence of observation  using this visualization. On the other hand, the robust kernel CCA has almost similar results to both data sets, ID and CD.
%here pic

\begin{figure*}
\begin{center}
\includegraphics[width=14cm, height=12cm]{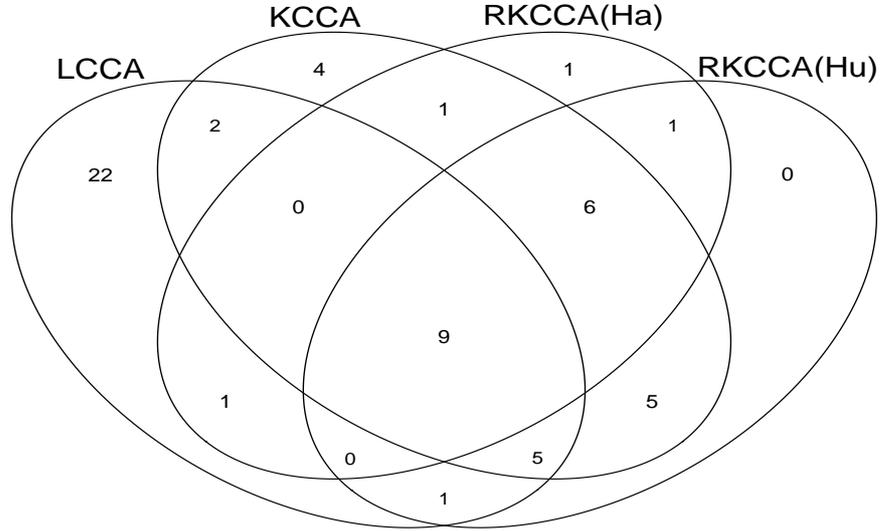}
\caption{Venn diagram of  significant  selected gene using  4 methods: LCCA, KCCA, RKCCA(Ha) and RKCCA(Ha)}
\label{ven}
\end{center}
\end{figure*}

\begin{table*}
 \begin{center}
\caption {The significant gene-gene co-associations based on  kernel CCA (first $14$ pairs out of $52$ pairs)}
\label{tb:sigLK}
 \begin{tabular}{ccccccccccc} \hline
\multicolumn{2}{c}{\rm{Gene}}&\multicolumn{2}{c}{\rm{LCCA}}&\multicolumn{2}{c}{\rm{KCCA}}\tabularnewline  
1&2&$T_{12}$&\rm{P-value}&$T_{12}$&\rm{P-value}\tabularnewline \hline
$\rm{GABRP}$&$\rm{HTR3A}$&$-0.383534$&$3.544411$&$0.701324$&$0.000393$\tabularnewline
$ \rm{BDNF}
$&$\rm{HTR2A}$&$ 0.813163$&$2.034120$&$0.416125$&$0.041939$\tabularnewline
$\rm{C1QTNF7}$&$\rm{MAGI2}
$&$ 1.000131$&$2.036340$&$0.317247$&$0.041716$\tabularnewline
$\rm{CHAT}$&$\rm{DRD2}$&$ 0.707203$&$2.033931$&$0.479441$&$0.041959$\tabularnewline
$\rm{CHAT}$&$\rm{ERBB4}$&$ 0.929514$&$2.013826$&$0.352623$&$0.044028$\tabularnewline
$\rm{CHAT}$&$\rm{GRM3}$&$ 0.163861$&$2.182260$&$0.869841$&$0.029090$\tabularnewline
$\rm{CHL1}$&$\rm{MAGI2}$&$ 0.880955$&$2.108948$&$0.378342$&$0.034949$\tabularnewline
$\rm{CHL1}$&$\rm{TPH1}$&$ 1.224253$&$2.250524$&$0.220857$&$0.024416$\tabularnewline
$ \rm{CLINT1}$&$\rm{COMTD1}$&$ 0.997200$&$2.126647$&$0.318668$&$0.033449$\tabularnewline
$\rm{CLINT1}$&$\rm{ERBB4}$&$ 1.020745$&$2.145911$&$0.307375$&$0.031880$\tabularnewline
$ \rm{CLINT1}$&$\rm{GRM3}$&$ 1.140133$&$1.994295$&$0.254231$&$0.046120$\tabularnewline
$ \rm{CLINT1}$&$\rm{GSK3B}$&$ 1.775388$&$2.251351$&$0.075834$&$0.024363$\tabularnewline
$\rm{COMTD1}$&$\rm{ERBB4}$&$ 1.209239$&$2.530227$&$0.226571$&$0.011399$\tabularnewline
$\rm{COMTD1}$&$\rm{FOXP2}$&$ 0.217550$&$2.110018$&$0.827780$&$0.034857$\tabularnewline
\hline
\end{tabular}
%}
\end{center}
\end{table*}

\begin{table*}
 \begin{center}
%\scalebox{0.55}[0.6]{
\caption {The significant gene-gene co-associations based on Robust  kernel CCA (RKCCA(Ha))}
\label{tb:Ha}
 \begin{tabular}{cccccccccccc} \hline
\multicolumn{2}{c}{\rm{Gene}}&\multicolumn{2}{c}{\rm{LCCA}}&\multicolumn{2}{c}{\rm{KCCA}}&\multicolumn{2}{c}{\rm{RKCCA(Hu)}}&\multicolumn{2}{c}{\rm{RKCCA(Ha)}}\tabularnewline  
1&2&$T_{12}$&\rm{P-value}&$T_{12}$&\rm{P-value}&$T_{12}$&\rm{P-value}&$T_{12}$&\rm{P-value}\tabularnewline  \hline 
\rm{BDNF}&\rm{MTHFR}&$1.1502$&$0.2501$&$2.0173$&$0.0437$&$2.0183$&$0.0436$&$1.8010$&$0.0417$\tabularnewline
\rm{CHL1}&\rm{NR4A2}&$0.6025$&$0.5469$&$1.5510$&$0.1209$&$1.8729$&$0.0611$&$2.0254$&$0.0428$\tabularnewline
\rm{CLINT1}&\rm{GSK3B}&$0.6676$&$0.5044$&$1.4336$&$0.1517$&$2.0040$&$0.0451$&$2.2394$&$0.0251$\tabularnewline
\rm{DTNBP1}&\rm{PIP4K2A}&$0.5164$&$0.6056$&$1.2807$&$0.2003$&$2.2737$&$0.0230$&$1.9912$&$0.0465$\tabularnewline
\rm{FOXP2}&\rm{GABRP}&$0.6119$&$0.5406$&$1.6017$&$0.1092$&$2.1579$&$0.0309$&$2.2054$&$0.0274$\tabularnewline
\rm{GABRA6}&\rm{GABRB2}&$1.0701$&$0.2846$&$1.2007$&$0.2299$&$1.8175$&$0.0691$&$2.0692$&$0.0385$\tabularnewline
\rm{GABRP}&\rm{PDLIM5}&$0.2299$&$0.8182$&$0.9791$&$0.3275$&$1.7081$&$0.0876$&$2.0207$&$0.0433$\tabularnewline
\rm{GRM3}&\rm{MAGI1}&$0.5145$&$0.6069$&$1.3759$&$0.1688$&$2.2942$&$0.0218$&$2.0027$&$0.0452$\tabularnewline
\rm{MAGI1}&\rm{NR4A2}&$0.5763$&$0.5644$&$1.5215$&$0.1281$&$2.2603$&$0.0238$&$1.9820$&$0.0475$\tabularnewline
\rm{MAGI2}&\rm{SLC1A2}&$1.8967$&$0.0579$&$0.9204$&$0.3573$&$2.1657$&$0.0303$&$2.4547$&$0.0141$\tabularnewline
\rm{MAGI2}&\rm{ST8SIA2}&$2.0275$&$0.0426$&$1.4559$&$0.1454$&$1.7892$&$0.0736$&$2.1058$&$0.0352$\tabularnewline
\rm{PDLIM5}&\rm{SLC1A2}&$0.7723$&$0.4400$&$1.4914$&$0.1359$&$1.8639$&$0.0623$&$2.1297$&$0.0332$\tabularnewline
\rm{SLC1A2}&\rm{ST8SIA2}&$0.9418$&$0.3463$&$1.2138$&$0.2248$&$2.0618$&$0.0392$&$2.1601$&$0.0308$\tabularnewline\hline 
\end{tabular}
%}
\end{center}
\end{table*}

\begin{table*}
\label{tab:sgene}
 \begin{center}
\caption {The selected genes using linear CCA (LCCA), kernel CCA (KCCA), robust kernel CCA.}
\scalebox{0.7}[0.8]{
 \begin{tabular}{lcccccccccc}\hline
\rm{Method}&\multicolumn{10}{c}{\rm{Gene}}  \\ \hline
& \rm{BDNF} & \rm{C1QTNF7}& \rm{CHGA} & \rm{  CLINT1} &\rm{ CSF2RA} &\rm{ DAO}  &\rm{ DGCR2}&\rm{DRD2} &\rm{DRD4} &\rm{ERBB3}  \\
&\rm{FOXP2}&\rm{FXYD2}  &\rm{GABBR1}&\rm{GABRB2}&\rm{GABRP} &\rm{GRIK3} &\rm{ GRM3} &\rm{GSTM1L}&\rm{HRH1}&\rm{HTR1A} \\
\rm{LCCA} &\rm{HTR2A}  &\rm{IL1B} &\rm{MAGI2}  &\rm{MICB} &\rm{MTHFR}&\rm{NOS1} &\rm{ NRG1}&\rm{NUMBL}&\rm{PDLIM5} &\rm{PLXNA2} \\
&\rm{PRODH2} &\rm{PTPRZ1} & \rm{RGS4} &\rm{SLC18A1} &\rm{SLC1A2} &\rm{SLC6A4} &\rm{ SOX10}&\rm{SRRM2}&\rm{SYNGR1}&\rm{TAAR6} \\\hline
 & \rm{BDNF}  &\rm{C1QTNF7}&\rm{CHAT} &\rm{CHL1} &\rm{CLINT1}&\rm{  COMTD1} &\rm{ DAOA}&\rm{DGCR2} &\rm{DISC1}&\rm{DRD2}\\
\rm{KCCA} &\rm{DTNBP1} &\rm{ERBB4} &\rm{FOXP2}  &\rm{GABRP} &\rm{GRIK3}&\rm{GRIN2B}&\rm{  GRM3} &\rm{GSK3B} &\rm{HRH1}&\rm{HTR2A} \\
  &\rm{HTR3A} &\rm{MAGI1}&\rm{MAGI2} & 
 \rm{MTHFR} &\rm{  NOTCH4}&\rm{NR4A2} &\rm{PDLIM5}&\rm{PIP4K2A}&\rm{PPP3CC} &\rm{ SLC1A2} \\
 &\rm{TAAR6} &  \rm{TPH1} &\\ \hline
&\rm{BDNF}&\rm{CHL1}&\rm{ CLINT1}&\rm{ DAOA}&\rm{DTNBP1}&\rm{ FOXP2}&\rm{GABRA6}&\rm{GABRB2}&\rm{GRM3}&\rm{GSK3B}\\
\rm{RKCCA(Ha)}&\rm{MAGI1} &\rm{MAGI2} &\rm{MTHFR}&\rm{NR4A2} &\rm{PDLIM5}& \rm{PIP4K2A}& \rm{SLC1A2}&\rm{ST8SIA2}&&\\ \hline
&\rm{BDNF}   & \rm{C1QTNF7} &\rm{CLINT1}  &\rm{COMTD1} &\rm{ CSF2RA}& \rm{ DAOA}& \rm{DISC1} &\rm{DRD2}  &\rm{DTNBP1}&\rm{ERBB4} \\
 \rm{RKCCA(Hu)}&\rm{FOXP2} &\rm{GABRP} &\rm{GRIK3}&\rm{GRIN2B}&\rm{GRM3}&\rm{GSK3B}&\rm{   HTR2A}&\rm{MAGI1}&\rm{MAGI2}&\rm{MTHFR}  \\
&\rm{NR4A2} &\rm{ PDLIM5}&\rm{PIP4K2A}&\rm{PPP3CC}&\rm{SLC1A2}&\rm{ST8SIA2}&\rm{ TAAR6}& && \\\hline
\end{tabular}
}
\end{center}
\end{table*}

\begin{table*}
 \begin{center}
\caption { A part of gene ontology and pathway analysis  of selected genes   using DAVID software of all methods}
\label{tb:pathway}
 \begin{tabular}{ccccccccccc} \hline
\rm{Database} &\rm{Term}&\rm{Method}&\rm{Count}&\%&\rm{P-Value}&\rm{Benjamini}\\\hline
& &\rm{LCCA}&$31$&$77.5$&$5.4E-32$&$4.2E-29$\\
& &\rm{KCCA}&$26$&$81.25$&$2.6E-25$&$1.2E-22$\\
& \rm{Schizophrenia(SZ)}&\rm{RKCCA(ha)}&$16$&$84.88$&$3.4E-16$&$9.9E-14$\\
\rm{GABDD}& &\rm{RKCCA(hu)}&$21$&$80$&$2.0E-21$&$8.7E-19$&\\\cline{2-7}
& &\rm{LCCA}&$5$&$12.8$&$1.1E-5$&$1.5E-3$\\
& &\rm{KCCA}&$6$&$18.8$&$4.4E-7$&$1.0E-7$\\
& \rm{SZ;bipolar disorder}&\rm{RKCCA(ha)}&$4$&$21.10$&$1.1E-4$&$1.6E-2$\\
& &\rm{RKCCA(hu)}&$21$&$80$&$2.0E-21$&$8.7E-19$&\\\hline
& &\rm{LCCA}&$11$&$28.4$&$3.5E-7$&$1.2E-5$\\
& &\rm{KCCA}&$8$&$25$&$2.3E-5$&$9.0E-4$\\
\rm{KEGG}& &\rm{RKCCA(ha)}&$4$&$21.1$&$1.2E-2$&$2.8E-1$\\
 &&\rm{RKCCA(hu)}&$7$&$2.7$&$8.0E-5$&$3.2E-3$&\\\hline
\end{tabular}
\end{center}
\end{table*}

\subsection{Mind Clinical Imaging Consortium (MCIC) Data analysis}
\label{Sec:RD}
Schizophrenia(SZ) is a  complex human  disorder that is  caused by the interplay of a  number of genetic  and environmental factors. The Mind Clinical Imaging Consortium (MCIC) has collected two types of data (SNPs and fMRI) from 208 subjects including $92$ SZ patients and $116$ healthy controls with $22442$ genes having $722177$  SNPs. Without missing data the number of subjects is $184$ ($81$ SZ patients and $103$ healthy controls) \citep{Dongdong-14}. For pairwise gene-gene interactions we consider top SZ $75$ genes, which are listed on the SZGene database (http://www.szgene.org). One gene does not have any SNPs. Finally,  we do the experiment on $74$ genes using linear CCA, kernel CCA and robust kernel CCA with two loss functions, which are described in Section $2$.  

We examine the linear CCA (LCCA), kernel CCA (KCCA) and robust kernel CCA with two functions Hample's and Huber's function, noted as RKCCA(Ha) and RKCCA(hu), respectively. In case of liner CCA and kernel CCA, on the one hand the liner CCA extracts $46$  significance  pairs with $40$ isolated genes, on the other hand the kernel CCA extracts  $52$ significance pairs with $32$ isolated genes at $5\%$ level of significance.  Table \ref{tb:sigLK} presents the first $14$ gene-gene co-associations based  on KCCA   along with values of test statistics and p-values. The robust methods, RKCCA(Ha)  and RKCCA(Hu)  extract $13$ and $33$ significance  pairs with $19$ and $27$ isolated genes, respectively. Table \ref{tb:Ha} shows $13$ pairs of  gene-gene co-association based RKCCA(Ha) along with values of test statistics and p-values. Table{tab:sgene} lists    all significance genes of linear CCA, kernel CCA and robust kernel CCA. To see the integration structure of the selected genes, we use the Venn-diagram of the four methods. Figure \ref{ven} presents the Venn-diagram of LCCA, KCCA, RKCCA(Ha) and RKCCA(Hu). By this figure we observe that the disjointly selected  genes of   LCCA, KCCA, RKCCA (Ha) and RKCCA (Hu)  are  $22$, $4$, $1$ and $0$. The number of common genes only between LCCA and KCCA, and LCCA and RKCCA, KCCA and RKCCA  are $16$, $2$  and $12$, respectively. All methods  select $9$ common  genes.

Finally, we conduct the gene ontology and the  pathway analysis using online software, the Database for Annotation, Visualization and Integrated Discovery (DAVID ) v6.7 \citep{DAVID}. The goal is to find the genes which are related to SZ disease. To do this, we consider the functional annotation chart of  DAVID. Table \ref{tb:pathway} consists of  a   part of the results of gene ontology and pathway analysis. This table contains count, percentages, adjusted P-values and Benjamini values of all methods. Note that the p-value is corrected for multiple hypothesis testing using the Benjamini-Hochberg method. In this table, GABDD and KEGG stand for Genetic association BD diseases and Kyoto encyclopedia of genes and genomes, respectively. On one hand, this table indicates that $84\%$  of the selected genes of the method RKCCA(Ha) are directly related to SZ diseases. On the other hand, those genes selected by linear CCA and kernel CCA are only $77.5\%$ and $81.25\%$ related, respectively. In case of  the term SZ and bipolar disorder RKCCA(Ha) gives better performance over all methods.

\section{Concluding remarks and future research}
\label{sec:con}
In this paper,  we have proposed   kernel CCU and its robust variants to detect gene-gene interaction  of SZ disease. The variances of kernel CCA, which is used in  kernel CCU, is estimated based on IF. In terms of ARE and  computational time, it is shown that  this estimator not only performs better in  ARE but also has a much lower computational time over bootstrap-based methods. The sensitivity analysis shows that the proposed robust kernel CCA is less sensitive to contamination than classical kernel CCA. We demonstrate the proposed methods to MCIC data set. Although the linear CCA and classical kernel CCA select a large set of  genes, these genes are less related to SZ disease.  On the other hand the robust methods are able to select a small set of  genes which are highly related to SZ disease. Based on  gene ontology and pathway analysis  we can conclude that the selected genes have  a significant influence on the manifestation of SZ disease.  

Although we illustrated the proposed methods only  to detect gene-gene interactions in SNPs data of MCIC, these  methods can also be extended to identify gene-gene interactions and ROI-ROI interactions  in DNA methylation data and fMRI data respectively.  The development of multiple kernel CCA based U statistics for use in more than two clinical trials in the future warrant valid inquiry for additional research.

\subsection*{Acknowledgments}
The authors wish to thank the NIH (R01 GM109068, R01 MH104680) and NSF (1539067) for support.

%\newpage
\begingroup
\bibliographystyle{plainnat}
\bibliography{Ref-UKIF}
\endgroup
%\newpage

\end{document}